\documentclass[letterpaper, 10 pt, conference]{ieeeconf}
\IEEEoverridecommandlockouts
\usepackage{amsthm}
\usepackage{amsmath,amssymb,amsfonts}
\usepackage{graphicx}
\usepackage{textcomp}
\usepackage{microtype}
\usepackage{booktabs}
\usepackage{tikz}
\usetikzlibrary{matrix,positioning,arrows.meta,calc}

\usepackage{amsmath,amsfonts,bm}

\usepackage{xspace}

\def\eqref#1{equation~\ref{#1}}

\def\1{\bm{1}}

\DeclareMathAlphabet{\mathsfit}{\encodingdefault}{\sfdefault}{m}{sl}
\SetMathAlphabet{\mathsfit}{bold}{\encodingdefault}{\sfdefault}{bx}{n}

\newcommand{\R}{\mathbb{R}}

\usepackage{float}
\usepackage{array}
\usepackage{subcaption}
\usepackage{pdflscape}
\usepackage{pdfpages}
\usepackage{wrapfig}
\usepackage{makecell}
\usepackage{multirow}

\usepackage{mine}

\usepackage{mathtools}
\usepackage{xcolor}

\newcommand{\todo}[1]{{\footnotesize \color{red} [TODO: {#1}]}}

\newcommand{\pmv}[2]{#1{\scriptstyle \pm #2}}
\definecolor{kpyellow}{HTML}{E8B800}
\definecolor{kpgreen}{HTML}{00AD09}
\newcommand{\swatch}[1]{\textcolor{#1}{$\bullet$}}

\newcommand{\draft}[1]{}

\newcommand{\averageFootDisplacement}{\bar{d}}
\newcommand{\averageGaitCycle}{\hat{f}}

\newcommand{\length}{L}
\newcommand{\height}{H}
\newcommand{\robotLength}{\length_r}
\newcommand{\robotHeight}{\height_r}
\newcommand{\animalLength}{\length_a}
\newcommand{\animalHeight}{\height_a}

\newcommand{\video}{\mathcal{V}}
\newcommand{\image}{\mathcal{I}}
\newcommand{\clipTime}{N}
\newcommand{\indexTime}{t}
\newcommand{\policy}{\boldsymbol{\pi}}

\newif\ifappendix
\appendixfalse   
\ifappendix
  \newcommand{\aref}[1]{~\hyperref[#1]{App.~\ref*{#1}}}
\else
  \newcommand{\aref}[1]{}
\fi
\begin{document}

\title{\LARGE \bf
\ourMethod: Quadrupedal Skill Transfer from a Single Side-View Video Without a Depth Model
}


\author{
Mohamad H. Danesh$^{*,1,2}$
Raina Hu$^{1}$
Mengxue Xu$^{1}$
Hsiu-Chin Lin$^{1,2}$
\thanks{*Corresponding author: {\tt\small mo.danesh@mail.mcgill.ca}}
\thanks{$^{1}$McGill University \quad $^{2}$Mila -- Quebec AI Institute}
}

\maketitle
\thispagestyle{empty}
\pagestyle{empty}

\begin{abstract}
Video-to-robot pipelines for quadrupedal imitation insert a learned lifting stage, either a monocular depth model or a per-clip 3D reconstruction, to recover the one axis a side-view camera cannot see. We present \ourMethod, which omits that stage and pins the missing axis to the robot's own stance width instead. The setting is side-view footage of sagittally dominated motion, where limb motion is almost entirely in the image plane. A frozen 2D keypoint detector and a fixed analytic retargeting then turn a single uncalibrated clip into a robot reference whose foot targets are reachable by construction, so no per-clip trajectory optimization is needed. One phase-indexed reward structure, whose only per-behavior freedom is three scalars from a shared sweep, learns a dog walk, a horse walk, a horse trot and two non-periodic gestures, all deployed zero-shot on hardware. We then measure what omitting the axis costs. In a training-free rigidity test, a video depth model varies the distance between the animal's hips by $40$ to $107\%$ of its own mean and a single-image metric model by $22$ to $58\%$, though that distance cannot move. Motion capture puts the true lateral paw travel of a dog walk at $4.5$ cm peak to peak, a mean deviation near $1.4$ cm against the $3.5$ cm of lateral foot error our policies already leave, so the discarded axis sits below the pipeline's resolution. Depth repaired to feasibility never beats the constant, and the unrepaired estimates put the robot on the ground.
\end{abstract}

\begin{figure*}[t]
\centering
\includegraphics[width=0.85\linewidth]{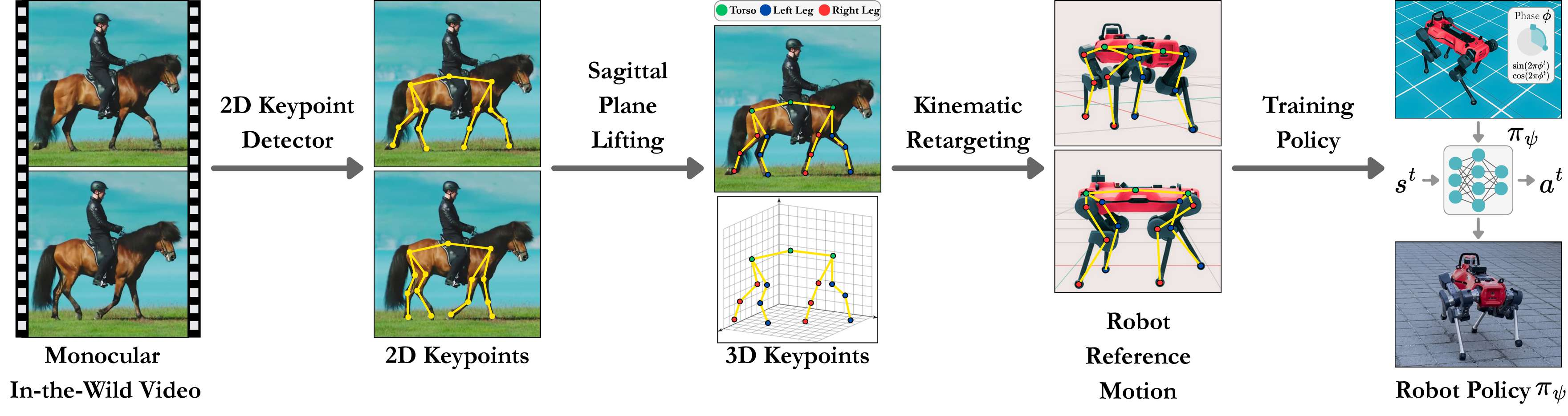}
\caption{\ourMethod. A single monocular clip becomes a robot-ready reference through one frozen 2D keypoint detector followed by a chain of fixed analytic steps, namely sagittal-plane lifting with the lateral axis pinned to the robot's stance width, kinematic retargeting, and root recovery. A phase-indexed policy $\policy_\psi$ trained in simulation tracks that reference and is deployed zero-shot on hardware. Only the reference itself and three reward scalars change between behaviors.} 
\label{fig:pipeline}
\end{figure*}

\begin{figure}[t]
\centering
\includegraphics[width=\columnwidth]{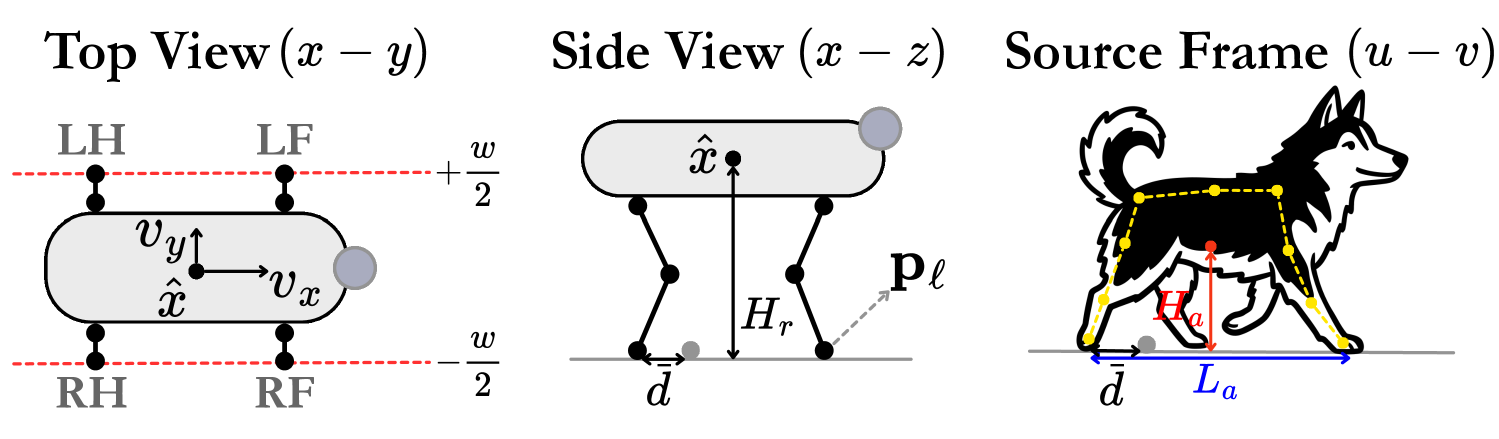}
\caption{
The reference quantities. \textbf{Left}: the stance tracks $y_k^{\text{nom}}{=}{\pm}w/2$ (\textcolor{red}{dashed}) that \autoref{eq:lift} pins the lateral axis to, with the base $\hat{\mathbf{x}}$ and speeds $v_x$, $v_y$. \textbf{Middle}: the robot height $H_r$, a foot $\mathbf{p}_\ell$ and the stance sweep $\bar d$. \textbf{Right}: the \textcolor[HTML]{bfaa0b}{detector keypoints}, all that is read from pixels, and the spans $L_a$, $H_a$ that \autoref{eq:scale} matches to $L_r$, $H_r$.
}
\label{fig:robot_frame}
\end{figure}

\section{Introduction}\label{sec:intro}

\noindent
Reinforcement learning (RL) has made quadrupeds robust \cite{lee2020learning,margolisyang2022rapid, rudin2022learning}, but producing animal-like motion is harder, since pure RL needs careful reward engineering and exploits simulation artifacts, yielding unnatural gaits that stress hardware~\cite{gurram2024rl}. Prior work therefore imitates motion capture (MoCap). Phase-indexed tracking reproduced dog MoCap on a real quadruped~\cite{peng2018deepmimic} and was later scaled to lifelike skill libraries~\cite{han2024lifelike}, but MoCap is expensive to collect, since it needs a dedicated laboratory and a trained animal.

Monocular video removes that cost and introduces a perception problem. A camera projects 3D space onto a 2D image, so depth is lost, and recovering it, a step known as \emph{lifting}, is ambiguous. Feet are also frequently occluded, so a naive lift yields a reference the robot cannot follow. Prior work therefore adds an explicit lifting stage, built either from species-specific 3D pose networks trained on scarce 3D ground truth~\cite{zhao2025learning,zhang2023slomo} or from per-clip trajectory optimization that repairs the lift into something feasible~\cite{yoon2024stmr}.

This paper asks whether that lifting stage is necessary for the most common kind of footage, a quadruped filmed from the side. Quadrupedal limb motion is predominantly parasagittal \cite{fischer2002basic,hildebrand1989quadrupedal}, so a side view shows almost everything a retargeter needs. The direction it hides, sideways across the body, is the one a depth model is called in to recover. We do not recover it. We pin it to the robot's own stance width and measure what that costs.

\ourMethod (\autoref{fig:pipeline}) turns an in-the-wild clip into a skill deployed zero-shot on a physical quadruped through one unmodified pipeline of a frozen 2D detector, a fixed analytic retargeting and one reward structure, whose only per-behavior freedom is three scalars from a small shared sweep. It learns a dog walk, a horse walk, a horse trot, a dog bow and a dog give-paw from one clip each.
We then evaluate the omitted axis three ways. A training-free \textit{rigidity test} asks the depth model to reproduce a distance that cannot move. A MoCap dog walk rendered to video decomposes the video reference against ground truth. Ablations substitute raw, metric and repaired depth for the constant. The claim is scoped to per-pixel monocular depth, since a per-clip articulated reconstruction does hold the rigid axis, but only by paying for a per-clip optimization, and it does not track any better than the pin (\autoref{subsec:exp-baselines}, \autoref{subsec:exp-minimal}).
Our key contributions are:
\begin{itemize}
\item A depth-free video-to-robot pipeline that turns a single uncalibrated side-view clip into a quadrupedal skill deployed zero-shot on hardware, with a frozen 2D detector, a fixed analytic retargeting and one reward structure across five behaviors and two species.
\item A retargeting rule that pins the lateral axis to the robot's own stance width. Reachable foot targets follow by construction rather than as a result, which is what lets the pipeline drop the per-clip trajectory optimization prior work runs to obtain them.
\item A measurement of what the omitted axis is worth, by a training-free rigidity test of monocular depth, a MoCap-versus-video decomposition of the reference error.
\end{itemize}

\section{Related Work}
\label{sec:related_work}

\noindent
\textbf{Motion retargeting.}
Animal and robot morphologies differ, so a reference motion must be retargeted rather than replayed. Prior work tracks the animal's foot positions and solves IK for joint targets the robot can reach~\cite{peng2020learning}, while SLoMo reconstructs a physically plausible 3D trajectory per clip and tracks it with trajectory optimization and model predictive control~\cite{zhang2023slomo}. 
\cite{peng2020learning} is the closest recipe to ours, and the difference is where it starts. It takes 3D keypoints as given, so on video it needs a lifting stage upstream and inherits what that stage gets wrong. We start from pixels and never form a 3D estimate, which is what makes the stage removable.

\textbf{Learning from video.} Humanoid imitation from monocular video leans on human-specific 3D priors \cite{videomimic,he2025asapaligningsimulationrealworld} that do not exist for arbitrary quadrupeds. VIAN \cite{yao2022imitation} keeps DeepLabCut \cite{mathis2018deeplabcut} keypoints in the image plane and lets the RL policy rescale them online under a per-behavior consistency reward, and RLWAV \cite{chane2024reinforcement} replaces the reference altogether with a video-classifier reward. We also read keypoints with a DeepLabCut-family detector \cite{ye2024superanimal}, but pair it with a fixed analytic retargeting, one reward structure carrying no per-behavior terms, and one clip per behavior.

\textbf{Lifting 3D animal motion.} Recovering 3D animal pose from one camera is hard, with ambiguous depth and scarce 3D annotation \cite{joska2021acinoset}. Prior work fits category-level shape models \cite{rueegg2023bite} and lifts 2D detections with task-specific networks \cite{zhao2025learning}. A separate line crosses the morphology gap, by conditioning on it \cite{danesh2026toward} or by per-clip optimization \cite{yoon2024stmr}, but starts from input that is already 3D. All treat an explicit 3D lifting stage as a prerequisite for retargeting, and none asks what that stage contributes. We ask exactly that, for side-view footage whose limb motion is already almost entirely in the image plane.

\section{\ourMethod}\label{sec:method}

\subsection{Problem Formulation}
\noindent
Given a monocular video clip $\video= \{ \image^\indexTime \}_{\indexTime=1}^{\clipTime}$ of $\clipTime$ frames, our goal is to learn a policy $\policy_\psi$ with parameters $\psi$ that reproduces the same behavior on a quadruped robot. The clip is the only demonstration, and the robot has never performed the behavior.

The target platform (\autoref{fig:pipeline}) is any quadruped with $12$ actuated joints $\mathbf{q}\in\R^{12}$, three per leg. Each leg carries one hip abduction/adduction joint (HAA) swinging the foot sideways, and hip and knee flexion/extension joints (HFE and KFE) moving it within the sagittal plane. This split matters throughout, because the camera sees the two sagittal joints and not the lateral one.
We write $\mathbf{p}_\ell\in\R^3$ for the position of foot $\ell\in\{\mathrm{LF},\mathrm{RF},\mathrm{LH},\mathrm{RH}\}$ in the base frame (\autoref{fig:robot_frame}), the two letters naming the left or right side and the front or hind pair. The pipeline needs only link lengths, stance width and joint limits, so it applies to any quadruped.

We assume \textbf{(i)} a static or slowly panning camera, \textbf{(ii)} travel roughly perpendicular to the optical axis, so the image plane is a weak-perspective projection of the sagittal plane, \textbf{(iii)} roughly constant subject depth, and \textbf{(iv)} steady-state behavior (near-constant-speed straight-line locomotion) or a gesture returning to its starting posture.



The output has two parts, a robot-executable reference trajectory of joint angles, base-frame foot positions and a world-frame base trajectory, and a closed-loop policy realizing it from proprioception alone. \ourMethod (\autoref{fig:pipeline}) produces both, with nothing in the video-to-reference chain trained or fitted per clip.

\begin{figure*}[t!]
\centering
\includegraphics[width=\linewidth]{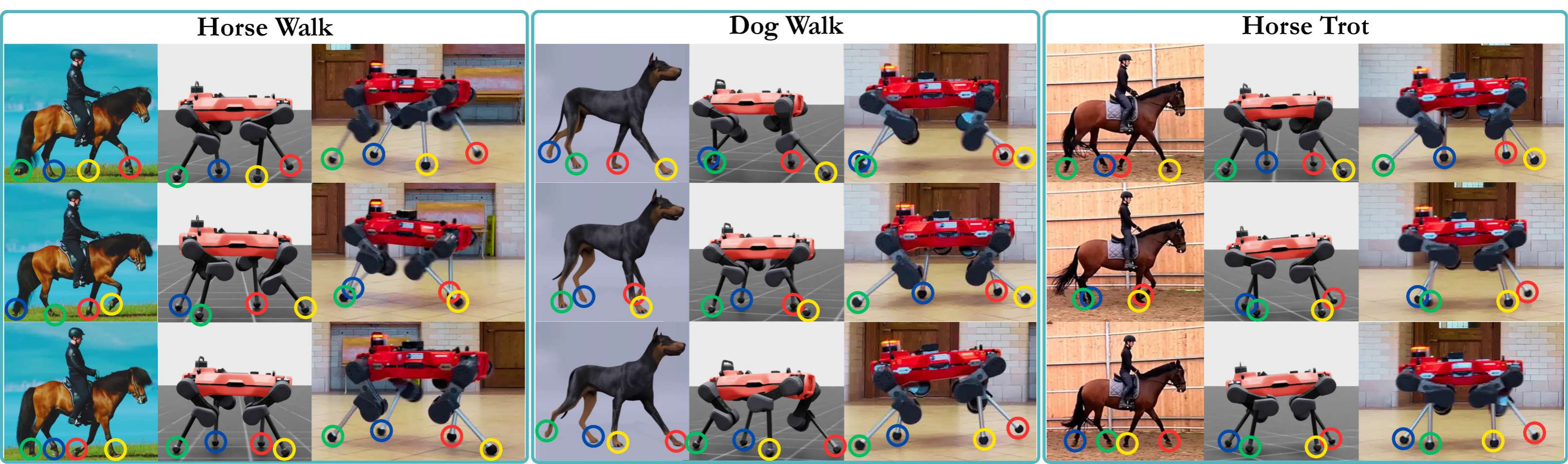}
\caption{Source video $\to$ Isaac Lab $\to$ real robot for the two gesture clips. All four feet are tracked, but only one front and one hind are ringed, \swatch{kpyellow} front and \swatch{kpgreen} hind, since left and right overlap in a side view.}
\label{fig:keyframes_1}
\end{figure*}

\subsection{2D Keypoints and Sagittal Lifting}\label{subsec:lift}
\noindent
We run a frozen off-the-shelf 2D keypoint detector~\cite{ye2024superanimal} on every frame zero-shot, without per-clip fine-tuning, and keep the $K$ of its outputs with an unambiguous counterpart on the robot. These are the paw, knee and hip of each limb, giving the foot targets, body orientation and leg configuration, plus three spine points (neck base, mid-back, tail base) defining the trunk axis. Throughout, $k$ indexes a detected keypoint and $\ell$ a robot foot.
Each keypoint $k$ in frame $\image^\indexTime$ yields a pixel location $(u_k^t,v_k^t)$ and a confidence $c_k^t\in[0,1]$, with $u_k^t$ the horizontal image coordinate aligned with the animal's direction of travel and $v_k^t$ the vertical one, measured downward from the top of the image. 


A side view supplies two of the three coordinates the retargeter needs, fore-aft and vertical, and hides the third. Rather than estimate the hidden one, we set it to a fixed value read off the robot's morphology:
\begin{equation}
\mathbf{p}_k^t = \big( s_x u_k^t,\ y_k^{\mathrm{nom}},\ -s_z v_k^t \big)
\label{eq:lift}
\end{equation}
where $y_k^{\mathrm{nom}}={\pm}w/2$ places left and right limbs on the robot's own stance tracks of width $w$ (\autoref{fig:robot_frame}, left) and $y_k^{\mathrm{nom}}{=}0$ puts the spine points on the midline. The vertical coordinate is negated because $v_k^t$ grows downward in the image while the robot's vertical axis points up.
The scalars $s_x$ and $s_z$ convert pixels to meters along the two visible axes. Both are fixed once per clip by matching the animal to the robot:
\begin{equation}
s_x = \frac{\robotLength}{\animalLength},
\qquad
s_z = \frac{\robotHeight}{\animalHeight}
\label{eq:scale}
\end{equation}
where $\animalLength$, $\animalHeight$ are the animal's fore-aft length (mean front-to-hind paw separation) and standing height (mean root-to-paw height) in pixels, averaged over the clip (\autoref{fig:robot_frame}, right), and $\robotLength$, $\robotHeight$ the robot's length and height at nominal stance.

The stance width sets the lateral scale directly, since \autoref{eq:lift} pins that axis. The match is purely geometric and not Froude-matched \cite{alexander1983dynamic}, and no camera calibration or depth estimate enters anywhere.

Two properties motivate this. Because the lateral coordinate is the robot's own, every foot target lies on a track the robot can reach, on every clip and with no per-clip repair. And a quadruped's lateral limb motion is small, so at the accuracy the rest of the pipeline reaches an estimated third coordinate contributes more noise than signal. The design therefore holds only while the true lateral excursion stays small, a boundary we locate on real gaits in \autoref{subsec:exp-mocap}.


%

\subsection{Kinematic Retargeting and Root Recovery}\label{subsec:retarget}
\noindent
The lifted keypoints $\mathbf{p}_k^t$ describe the animal's body configuration, but the robot needs joint angles and a base trajectory, so we map them to a feasible reference $\{\hat{\mathbf{q}}^t,\hat{\mathbf{p}}_\ell^t,\hat{\mathbf{x}}^t\}$ of joint angles, base-frame foot positions and base position. A hat marks a quantity derived from the video throughout, as opposed to one the robot achieves. We call the origin of the base frame the root, and the stage recovering its trajectory \textit{root recovery}.

\textbf{Preprocessing.} 
Each keypoint track is cleaned of detector noise independently. Frames with confidence below $c_{\min}$ are linearly interpolated per coordinate from the nearest frames on either side that clear $c_{\min}$, and the track is then passed through a centered moving-average filter, edge-padded so the clip's first and last frames survive.

\textbf{Hip regularization.}
The four hip keypoints are pulled toward a rigid rectangle by a quadratic penalty of weight $\lambda_h$ that shrinks each hip toward its nominal position in the body frame, which stabilizes the estimated body heading. The base position is set to the center of the trunk at the mean hip height.

\textbf{Joint positions through inverse kinematics.} The steps so far give scaled foot targets, not joint angles. Holding the base fixed, we recover the latter with damped least-squares differential IK:
\begin{equation}
\hat{\mathbf{q}}_\ell \leftarrow \Pi\!\left[\hat{\mathbf{q}}_\ell + \eta \mathbf{J}_\ell^\top \big(\mathbf{J}_\ell\mathbf{J}_\ell^\top+\lambda_I^2\mathbf{I}\big)^{-1}\big(\hat{\mathbf{p}}_\ell-\mathbf{p}_\ell(\hat{\mathbf{q}}_\ell)\big)\right]
\label{eq:ik}
\end{equation}
per leg $\ell$ over a fixed per-frame iteration budget, where $\hat{\mathbf{q}}_\ell\in\R^3$ collects that leg's three reference joint angles, $\mathbf{p}_\ell(\cdot)$ is its forward kinematics, $\mathbf{J}_\ell$ the foot Jacobian, $\eta$ a step size, $\lambda_I$ a Tikhonov damping term that trades exactness for conditioning near singular leg configurations, and $\Pi[\cdot]$ the projection onto the robot's soft joint limits. Reachability is enforced one level earlier. Each target $\hat{\mathbf{p}}_\ell$ is confined to a shell inside the leg's workspace before solving, so the solver never drives joints toward their stops (verified on all emitted references). This clamp is one fixed rule applied identically to every clip, on the two axes the camera sees, and with the pin it replaces per-clip trajectory optimization.

\textbf{Root recovery.}
\label{sec:base-recovery}
Because IK solves for joint angles with the base held fixed, the animal's forward translation is not captured in $\hat{\mathbf{q}}$.
To recover it, we estimate two quantities, the foot sweep per gait cycle $\averageFootDisplacement$ in meters (\autoref{fig:robot_frame}, middle) and the gait frequency $\averageGaitCycle$ in cycles per second.

For the foot sweep we use the fact that while a foot is planted, its backward motion in the body frame equals the body's forward motion in the world frame. We take the peak-to-peak fore-aft excursion of each foot in the body frame and set $\averageFootDisplacement$ to the median over the four feet, which is robust to the one or two tracks the detector loses. For the frequency, each time a foot lifts and reaches its highest point counts as one footfall, and $\averageGaitCycle$ is the median footfall count over the four feet divided by the clip duration. The forward speed is then:
\begin{equation}
\hat{v} = \averageFootDisplacement \times \averageGaitCycle
\label{equ:forward-speed}
\end{equation}
When a clip holds under two gait cycles, the cadence estimate is unreliable and we fall back to the per-frame median backward velocity of the currently planted feet. Integrating the constant $\hat v$ gives a straight-line base trajectory $\hat{\mathbf{x}}^t$ at constant heading, which supplies the three root targets the policy tracks, the forward speed $\hat v_x{=}\hat v$, the base height $\hat h^t$, and a zero lateral velocity, since the pin admits no sideways translation. For the three gaits $\hat h^t$ is the constant mean scaled root height, and for the two gestures it follows the scaled root height frame by frame, which is what carries the crouch of the bow. Within-stride surge, heave and sway are not recovered.

\subsection{Phase-Indexed Imitation Policy}\label{subsec:policy}
\noindent
We train one policy per behavior with a phase-indexed DeepMimic formulation \cite{peng2018deepmimic}. The reference is a kinematic target rather than a replayable trajectory, and what is learned is a closed-loop controller that reproduces it while keeping the robot balanced under its own dynamics and contact. From each clip we take the steady-state segment of duration $T$ and define the phase $\phi^t=(t \bmod T)/T$. The policy observes a state $\mathbf{s}^t$ of $48$ proprioceptive features (joint positions and velocities, base orientation and angular velocity, previous action, velocity command) plus $(\sin2\pi\phi^t,\cos2\pi\phi^t)$, and outputs an action $\mathbf{a}^t\in\R^{12}$ of joint-angle offsets from the nominal stance, converted to joint torques.

\begin{figure}[t]
\centering
\includegraphics[width=\linewidth]{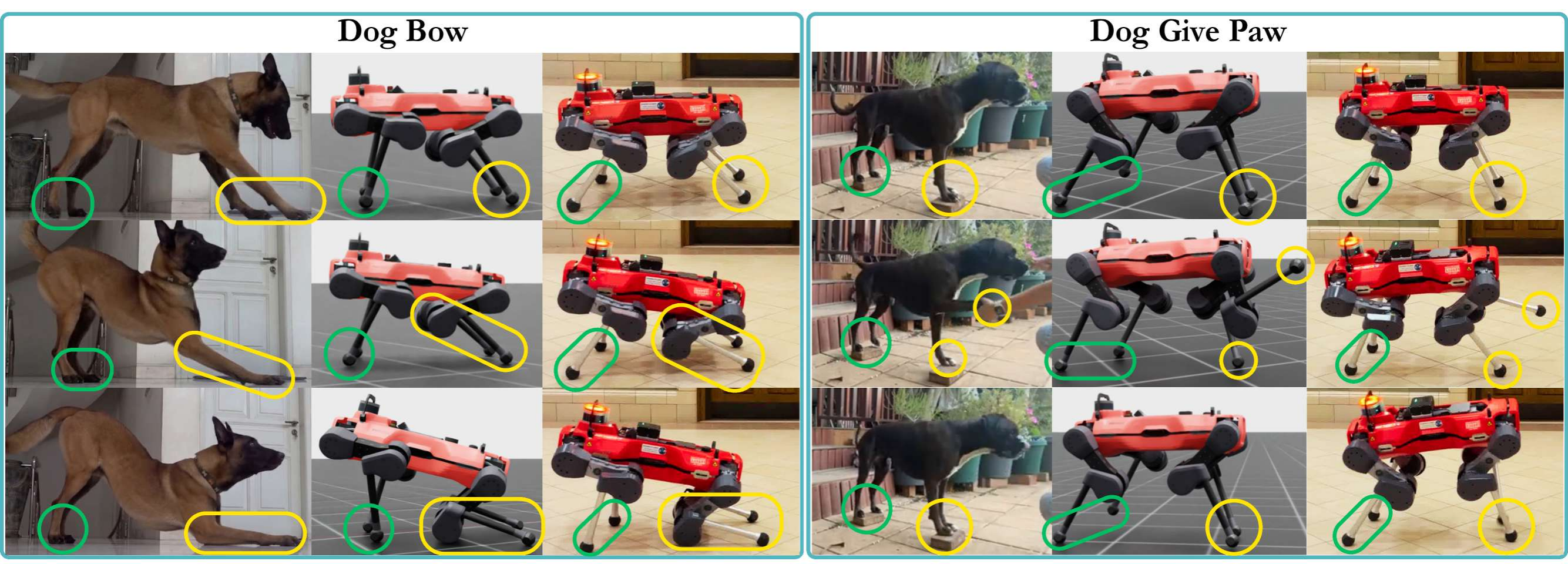}
\caption{Source video $\to$ Isaac Lab $\to$ real robot for the two gesture clips. The gestures are read from the near-side pair alone, so only two feet are ringed, \swatch{kpyellow} front and \swatch{kpgreen} hind.}
\label{fig:keyframes_2}
\end{figure}

\textbf{Reward.} The policy is rewarded for matching the reference at the current phase:
\begin{equation}
r^t = w_j r_j^t + w_f r_f^t + w_r r_r^t - w_a r_a^t
\label{eq:reward}
\end{equation}

\noindent
Each reward term is defined as:
\begingroup
\begin{align}
r_j^t &= \exp \big(-\|\mathbf{q}^t-\hat{\mathbf{q}}^t\|^2 \big/ \sigma_j^2 \big) 
\nonumber \\
r_f^t &= \exp \Big(-\textstyle\sum_\ell
\|\mathbf{W}_f^{1/2}\big(\mathbf{p}_\ell^t-\hat{\mathbf{p}}_\ell^t\big)\|^2 \big/ \sigma_f^2\Big) \nonumber\\
r_r^t &= \exp \Big(-\big[
(v_x^t-\hat v_x)^2 + (v_y^t)^2 +
(h^t-\hat h^t)^2
\big] \big/ \sigma_r^2\Big) 
\nonumber\\
r_a^t &= \|\mathbf{a}^t-\mathbf{a}^{t-1}\|^2
\label{eq:rr}
\end{align}
\endgroup

\noindent
where
$r_j$ rewards agreement with the reference joint angle.
$r_f$ rewards agreement with the reference feet positions $\mathbf{p}_\ell^t$, $\mathbf{W}_f = \mathrm{diag}(1,1,w_z)$, and $w_z$ sets how much the vertical component counts against the two horizontal ones. 
$r_r$ rewards agreement with the recovered root motion of \autoref{subsec:retarget}, where $v_x^t$, $v_y^t$ and $h^t$ are the robot's achieved forward speed, lateral speed and base height, measured against the targets $\hat v_x$, zero and $\hat h^t$. It has no orientation component, and base pitch is never commanded.
$r_a$ penalizes the action rate, which suppresses chatter that hardware cannot execute.

The $w$'s weight the four terms and the $\sigma$'s set each exponential's tolerance. The same reward is used for all five behaviors, the bow and give-paw entering only through the joint reference and the time-varying height target $\hat h^t$, with $\hat v_x{=}0$. The only per-behavior freedom is the three scalars $w_r$, $w_a$, $w_z$, selected per clip from a shared $3{\times}3{\times}3$ grid.

\begin{table*}[t!]
\centering
\caption{All five behaviors on ANYmal-D, from the source clip through simulation to zero-shot hardware. \textit{$e_q$ per-clip} and \textit{$e_q$ shared} differ only in the reward weights, namely three per-clip scalars or one fixed vector for all five, and both are evaluated against the same full-method reference, so the per-clip row is the \textit{full} bar of \autoref{fig:abl}. A dash marks a quantity undefined for a stationary gesture.
}
\label{tab:all}
\begin{tabular*}{\linewidth}{@{\extracolsep{\fill}}llccccc@{}}
\toprule
Source clip & & Dog walk & Horse walk & Horse trot & Dog bow & Dog give-paw \\
\midrule
 & fps, length (s) & 30, 21.4 & 25, 21 & 30, 36 & 50, 7 & 30, 8 \\
\midrule
\multirow{6}{2cm}{\textit{Simulation}}
& $\hat{\bar v}\to\bar v$ (m/s) & $0.29\to0.29$ & $0.48\to0.55$ & $0.58\to0.58$ & $0.00\to0.00$ & $0.00\to0.00$ \\
& $e_q$ per-clip (deg)$\downarrow$ & $\pmv{14.3}{1.4}$ & $\pmv{13.3}{1.7}$ & $\pmv{11.6}{1.3}$ & $\pmv{4.7}{0.3}$ & $\pmv{8.3}{4.4}$ \\
& $e_q$ shared (deg)$\downarrow$ & $\pmv{18.1}{1.6}$ & $\pmv{15.7}{2.1}$ & $\pmv{13.4}{1.9}$ & $\pmv{4.2}{0.4}$ & $\pmv{4.6}{0.7}$ \\
& $e_p$ (cm)$\downarrow$ & $\pmv{13.7}{1.8}$ & $\pmv{15.0}{2.6}$ & $\pmv{15.5}{2.0}$ & $\pmv{7.4}{0.4}$ & $\pmv{11.8}{6.8}$ \\
& $e_p$ fore-aft / lateral / vert.\ (cm) & $9.8/3.5/3.7$ & $10.5/3.8/4.0$ & $11.0/3.9/4.2$ & --- & --- \\
& Surv.\ (s, cap 20) & $\pmv{19.9}{0.2}$ & $\pmv{19.5}{0.7}$ & $\pmv{19.6}{0.5}$ & $\pmv{20.0}{0.0}$ & $\pmv{15.1}{2.9}$ \\
\midrule
\multirow{4}{2cm}{\textit{Hardware, zero-shot}}
& $\hat{\bar v}\to\bar v$ (m/s) & $0.29\to\pmv{0.26}{0.02}$ & $0.48\to\pmv{0.49}{0.02}$ & $0.58\to\pmv{0.52}{0.03}$ & $0.00\to0.00$ & $0.00\to0.00$ \\
& $e_q$ (deg)$\downarrow$ & $\pmv{16.8}{1.1}$ & $\pmv{15.4}{1.3}$ & $\pmv{13.5}{1.0}$ & $\pmv{5.9}{0.5}$ & $\pmv{9.6}{2.8}$ \\
& $e_p$ (cm)$\downarrow$ & $\pmv{15.9}{1.4}$ & $\pmv{17.1}{1.7}$ & $\pmv{17.8}{1.5}$ & $\pmv{8.6}{0.6}$ & $\pmv{13.1}{3.9}$ \\
& Surv.\ (s, cap 20) & $\pmv{20.0}{0.0}$ & $\pmv{20.0}{0.0}$ & $\pmv{20.0}{0.0}$ & $\pmv{20.0}{0.0}$ & $\pmv{19.0}{2.0}$ \\
\bottomrule
\end{tabular*}
\end{table*}

\section{Experiments}\label{sec:experiments}
\noindent
We ask: \textbf{(i)} Does one unmodified pipeline reproduce distinct behaviors in simulation and on hardware (\autoref{subsec:exp-sim})? \textbf{(ii)} How does the method compare against alternatives that replace the lifter or the objective (\autoref{subsec:exp-baselines})? \textbf{(iii)} Is the depth axis worth recovering (\autoref{subsec:exp-minimal})? \textbf{(iv)} Which of the remaining components are necessary (\autoref{subsec:exp-root}, \autoref{subsec:exp-abl})? \textbf{(v)} What does video cost against MoCap (\autoref{subsec:exp-mocap})? \textbf{(vi)} Does the pipeline hold beyond the five clips (\autoref{subsec:exp-breadth})?

\subsection{Experimental Setup}\label{subsec:exp-setup}
\noindent
\textbf{Clips.} Each behavior is learned from a \textit{single} uncalibrated internet clip. Of $24$ candidate clips screened against the four assumptions of \autoref{sec:method}, $14$ passed, so close to three in five casually chosen side-view clips are usable with no curation beyond the screen itself. The screen rejects oblique viewpoint, fast camera translation and near-side occlusion beyond what interpolation recovers.
All $14$ then produced trainable references whose joint angles clear the robot's soft limits with zero violations (worst-case margin $13^\circ$), the reachability guarantee of \autoref{subsec:lift} holding on every clip we tried.
\autoref{subsec:exp-sim} reports five of them, a dog walk, a horse walk, a horse trot, a dog bow and a dog give-paw (\autoref{fig:keyframes_1}, \autoref{fig:keyframes_2}), and \autoref{subsec:exp-breadth} the remaining nine.
 
\textbf{Training.} Actor and critic are MLPs ($512$-$256$-$128$ and $256$-$128$, ELU), the policy runs at $50$ Hz over a $200$ Hz physics step, and episodes last $20$ s. All policies train on ANYmal-D in Isaac Lab \cite{mittal2025isaac} on flat terrain under the same domain randomization (DR), namely static ground friction in $[0.6,1.0]$, base mass $\pm5$ kg and center of mass, actuator effort limit scaled by $[0.65,1.0]$, and mild external pushes, with no observation noise and no actuation latency. The policy is the one component learned from data, trained with PPO \cite{schulman2017ppo} over $512$ parallel environments \cite{rudin2022learning} using reference-state initialization at a random phase and early termination when the tracked pose deviates beyond a threshold. Training runs $3000-4000$ iterations of $512\times24$ steps (${\approx}37-49$M environment steps), GAE ($\gamma{=}0.99$, $\lambda{=}0.95$), clip ratio $0.2$, $4$ epochs per iteration at minibatch $512$, and a KL-adaptive learning rate initialized at $10^{-4}$. 

We use $K{=}15$ keypoints: the four paws, four knees, four hips and three spine points among SuperAnimal-Quadruped's $39$ outputs that have an unambiguous correspondence to ANYmal-D. The retargeter uses $c_{\min}{=}0.5$, a centered $7$-frame moving average over each keypoint track (${\approx}0.23$ s at $30$ fps, short enough to leave touchdown timing intact) and $\lambda_h{=}5$. IK uses $\lambda_I{=}0.05$, $\eta{=}1$ and $20$ iterations per frame. Shared reward settings are $\sigma_j{=}0.5$, $\sigma_f{=}0.25$, $\sigma_r{=}0.25$, $w_j{=}0.5$, $w_f{=}0.3$, a $2.0$ rad termination threshold, and the per-clip grid $w_r\in\{0.1,0.15,0.2\}$, $w_a\in\{0.2,0.3,0.4\}$, $w_z\in\{4,8,12\}$.

\textbf{Metrics.} Unless stated otherwise, every configuration is trained with $10$ random seeds and evaluated over $10$ episodes capped at $20$ s, and every hardware behavior is deployed for $10$ trials. We report Root Mean Square (RMS) per-joint angle error $e_q$ (deg) and mean per-foot Cartesian error $e_p$ (cm), evaluated from phase zero against the retargeted reference at the same phase, from simulator state and, on hardware, from encoder logs and leg forward kinematics. That reference is the target \autoref{eq:reward} asked the controller to reproduce rather than an independently measured ground truth, since an internet clip admits none. The one exception is the rendered MoCap sequence of \autoref{subsec:exp-mocap}, evaluated against the marker-derived reference. The error $e_q$ is taken jointly over the twelve joints, all environments and all timesteps, and accumulates through early termination, so a large $e_q$ indicates a fall rather than proportionally worse tracking. For ablations, $e_q^{\mathrm{ref}}$ evaluates every variant against the \textit{full} method's reference, so all variants share one fixed target. We also report achieved speed $\bar v$ against the target $\hat{\bar v}$ and survival. As a reference-free floor we evaluate a \textit{do-nothing} policy holding the nominal stance for the whole episode, which reaches $28-30^\circ$ on the three gaits. Values after $\pm$ are one standard deviation over seeds in simulation and over trials on hardware.

\subsection{Five Behaviors, in Simulation and on Hardware}\label{subsec:exp-sim}
\noindent
Using one pipeline and one reward structure, \ourMethod learns all five behaviors in simulation and transfers each to the physical quadruped zero-shot. The numerical outcomes are summarized in \autoref{tab:all}.

\textbf{Simulation.} 
The three gaits hit their recovered speeds to within $0.06$ m/s. Absolute agreement with the reference is coarse (\autoref{tab:all}), roughly half the $28-30^\circ$ of the do-nothing floor. The per-joint error concentrates in the sagittal hip flexion and knee (up to $15.5^\circ$), the joints driven from the image, and not in the pinned hip abduction. We therefore claim \textit{gait-structure} transfer rather than pose-level fidelity, and the keyframes of \autoref{fig:keyframes_1} show that structure surviving intact. The bow reproduces its stand$\to$bow$\to$stand cycle with a peak base-pitch error of $6.4\pm0.7^\circ$ on an axis that is never commanded, so that agreement emerges from tracking the legs alone. The give-paw balances on three legs while one paw rises and is barely statically stable, which is why it is the one behavior whose seeds differ. Both gestures occlude the far-side front limb. Its knee and paw fall below $c_{\min}$ on $15\%$ of bow frames, and on $66\%$ and $44\%$ of give-paw frames, where the raised limb is the far one. Those frames are interpolated (\autoref{subsec:retarget}) rather than mirrored or filled from a prior, and the pin leaves interpolation only the two visible axes. Every other limb stays above $0.7$ mean confidence. Our protocol deploys every seed rather than selecting a good one.

\textbf{Hardware.} We deploy every behavior to the physical ANYmal-D zero-shot, without fine-tuning, real-world data or gain retuning (see the supplementary video). Policies hold their commanded speeds to a ${\sim}10\%$ undershoot from simulation, consistent with foot slip. The phasing that defines each gait survives the transfer as well, visible in the hardware column of \autoref{fig:keyframes_1}. Survival matches or exceeds simulation for every behavior, because DR occasionally draws a simulated episode into an adverse corner, such as low friction or a weak actuator, that the single physical robot at its own fixed parameters never encounters. The marginal give-paw shows this most clearly, surviving \textit{longer} on the robot because its one failing seed tips over only under a weak actuator draw the robot never samples (narrowing that DR range to $[0.85,1.0]$ raises simulated survival to $18.7\pm1.4$ s), a case of hand-set ranges being wider than the hardware requires, which uncertainty-gated randomization addresses directly \cite{danesh2025safe}. 
Transfer succeeds with no observation noise or latency in the DR, which we attribute to $50$ Hz position targets and coarse tracking tolerances.

\draft{
Duty factor drifts upward at every step while phasing does not. Under both contact detectors it rises monotonically from reference to simulation to hardware, but the height detector rises only about half as far as the force detector, so roughly half the drift is the detector and half is real. Diagonal offset does not drift at all (\autoref{tab:all}). Phase structure is therefore what transfers, at a slower and more conservative operating point. Reference agreement on the robot degrades by $15-25\%$ from simulation across the five behaviors (\autoref{tab:all}). The gap is largest on the two walks, where the longer real stance phase lags the reference's swing timing, and smallest on the slow gestures.
}

\textbf{Keyframes.} \autoref{fig:keyframes_1} and \autoref{fig:keyframes_2} place the source frame, the simulated policy and the hardware rollout side by side at three matched phases per behavior, with the tracked feet ringed. Read left to right, each triplet reaches the animal's limb configuration at the same point in the cycle, in simulation and on the robot alike, and the gait pattern is legible directly from the ringed feet, with the trot's diagonal legs swinging and landing together while the walks move one leg at a time. The gestures are recognizable from posture alone, which is worth noting because base orientation is never commanded (\autoref{eq:rr}), so that posture comes entirely from tracking the legs.

\begin{table}[t]
\centering
\setlength{\tabcolsep}{3pt}
\caption{Alternatives on the dog walk. Each row replaces one stage and leaves the rest untouched, the upper group the lifting stage and the lower group the objective. All use $e_q^{\mathrm{ref}}$ against the full method's reference, survival capped at $20$ s. Dashes are undefined quantities. $^\dagger$ marks the need to an extra conversion (SLoMo) or a data regime we do not supply (AMP).
}
\label{tab:baselines}
\begin{tabular*}{\linewidth}{@{\extracolsep{\fill}}lccc@{}}
\toprule
& $e_q^{\mathrm{ref}}$ (deg)$\downarrow$ & $\bar v$ (m/s) & Surv.\ (s) \\
\midrule\midrule
\textbf{\ourMethod (ours)} & $\pmv{14.3}{1.4}$ & $0.29$ & $\pmv{19.9}{0.2}$ \\
\midrule\midrule
\multicolumn{4}{@{}c}{\textbf{Replacing the lifting stage}}\\
\midrule
SLoMo~\cite{zhang2023slomo}$^\dagger$ & $33.6-36.6$ & ${\approx}0$ & --- \\
SLoMo+physics refinement~\cite{zhang2023slomo,yang2023lab4d}$^\dagger$ & $\pmv{19.5}{1.1}$ & $\pmv{0.22}{0.1}$ & $20.0$ \\
\midrule\midrule
\multicolumn{4}{@{}c}{\textbf{Replacing the objective}}\\
\midrule
AMP \cite{peng2021amp}$^\dagger$ & --- & $0.03$ & --- \\
Velocity command only & ${\approx}28-30$ & $0.27$ & $20.0$ \\
\bottomrule
\end{tabular*}
\end{table}

\subsection{Comparison Against Alternative Lifters and Objectives}\label{subsec:exp-baselines}
\noindent
No prior system shares our pipeline, so a head-to-head run would confound every stage at once. We instead swap one stage at a time and hold the rest fixed.

\textbf{Replacing the lifting stage.}
SLoMo~\cite{zhang2023slomo} is a complete pipeline rather than a lifting stage. It reconstructs a per-clip 3D trajectory and tracks it with model predictive control, so using it as a lifter needs a conversion. We run its reconstruction~\cite{yang2023lab4d} on our dog-walk clip, map the recovered skeleton onto our keypoint interface and leave the rest unchanged, once raw (SLoMo) and once with the physics-informed refinement the original relies on (SLoMo+physics refinement).

The upper group of \autoref{tab:baselines} gives the result. The raw reconstruction damps articulation to a fore-aft foot sweep of $0.17$ stance lengths against $0.80$ for ours, so the trained policy stands rather than walks. Refinement recovers a walking gait, but stays $5^\circ$ above ours on the same target and needs a per-clip reconstruction and optimization to get there, where our lifting stage is a constant. The comparison runs on one clip through our conversion, which leaves an IK residual of $19-26$ cm where our references leave zero, so we read it as evidence on cost and feasibility rather than a fidelity verdict on learned reconstruction.


\draft{
Two rows cannot be reduced to a single substitution. We mark them and say what they do and do not license, rather than read them as head-to-head results.
Two findings then separate. The reconstruction's own geometry can be measured without our retargeter. It holds the rigid axis, \autoref{subsec:exp-minimal}, but its articulation is damped, with a fore-aft foot sweep of $0.17$ stance lengths against $0.80$ for ours, and the unrefined reference trains to a policy that stands rather than locomotes. The downstream policy numbers cannot be measured that way. Our conversion leaves an IK residual of $19-26$ cm where our own references leave zero, so the remaining gap in \autoref{tab:baselines} belongs to the reconstruction and to our conversion together. These rows are evidence on cost and per-clip feasibility, not a fidelity verdict on learned reconstruction.
}

\textbf{Replacing the training objective.}
We next keep our reference and replace \autoref{eq:reward}, holding observations, actions, actuator model and DR fixed. The two substitutions are
(1) AMP~\cite{peng2021amp}, which replaces phase-indexed tracking with an adversarial motion prior, and
(2) velocity command only, which deletes the imitation terms and leaves the root term.

The lower group of \autoref{tab:baselines} gives the result. AMP is built to match a distribution drawn from a large \textit{library} of clips rather than the single short clip our setting supplies. With one clip the discriminator separates reference from policy almost immediately and returns little usable gradient, so the policy never reaches locomotion ($0.03$ against $0.29$ m/s). This bounds AMP in the single-clip regime, not where its data assumption holds. ``Velocity command only'' does reach the target speed and survive, which is what a velocity reward is for, but from a crouch at $0.3$ m base height against the reference's $0.5$ m, with pose error at the do-nothing floor. Speed and survival come from the root term, the upright animal-like posture from the imitation terms.



\textbf{Methods not substituted.}
Two lines of work cannot be dropped into a single stage. \cite{zhao2025learning} trains a species-specific 3D pose network on 3D ground truth our setting does not supply and releases no code, so a substitution would be our reimplementation rather than the method. VIAN \cite{yao2022imitation} has no separable lifting stage at all, since it keeps its 2D keypoints in the image plane and lets the policy rescale them online under a per-behavior consistency reward, so swapping it in would replace lifter and objective at once.


\subsection{Is the Depth Axis Worth Recovering?}\label{subsec:exp-minimal}
\noindent
Our lifting method in \autoref{eq:lift} fixes the lateral axis to a constant $y_k^{\mathrm{nom}}{=}{\pm}w/2$.
Because prior pipelines repair the depth estimate through per-clip optimization, we test whether replacing \autoref{eq:lift} with a depth estimate improves tracking. Each variant changes only the lateral coordinate.

\textbf{Depth lifting at three levels of repair.}
We substitute three lateral estimates of increasing effort. ``Raw disparity'' takes the depth model's output directly. ``Metric depth'' converts that output from model units to world coordinates. ``Repaired depth'' additionally clamps the metric estimate frame by frame to the reachable workspace and low-pass filters it, which is what per-clip optimization buys in prior work.

\begin{figure}[t]
\centering
\includegraphics[width=0.8\linewidth]{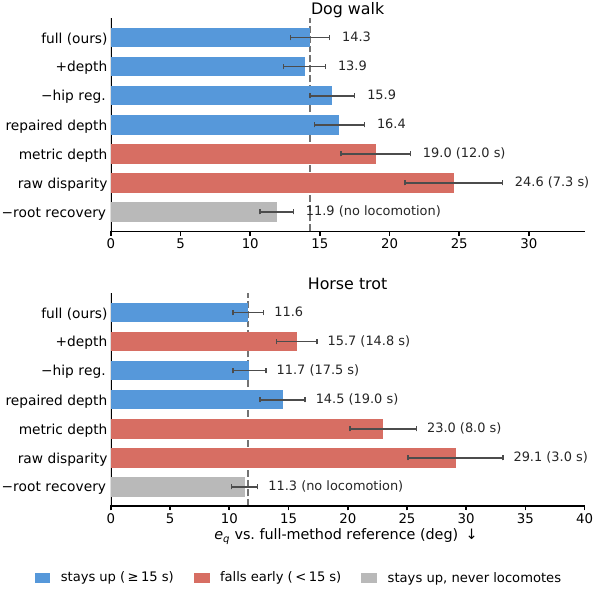}
\caption{Retargeting ablation on the dog walk and the horse trot. Bars are $e_q^{\mathrm{ref}}$ against the full method's reference. Every variant is evaluated on one fixed target, with whiskers $\pm1$ std over seeds. Color marks what the policy did, not how well it tracked. Blue survives the horizon and locomotes. Red falls before $15$ s, so its $e_q$ is post-fall accumulation rather than tracking, with survival annotated. Gray stays up but never locomotes, so a low $e_q$ is a stationary robot.
}
\label{fig:abl}
\end{figure}

The result is monotone in effort and never reaches the constant (\autoref{fig:abl}). Raw disparity puts the robot on the ground within seconds on both clips. Metric depth survives longer but still falls, because the hip separation it implies drifts between frames. Repair stops the falls, but the error stays $2-3^\circ$ above the pin on both clips ($16.4$ against $14.3$ on the dog walk, $14.5$ against $11.6$ on the trot). Three stages of machinery end up behind a single number read off the robot's own body.

\autoref{fig:abl} carries two further bars. ``+depth'' adds a monocular depth channel back upstream of the pin, moving the dog walk by $0.4^\circ$, inside one seed standard deviation, while the trot loses its footing and falls at $14.8$ s. ``-hip reg.'' drops the hip regularization of \autoref{subsec:retarget}, costing $1.6^\circ$ on the dog walk and leaving the trot unchanged.


\begin{table}[t]
\centering
\setlength{\tabcolsep}{2.2pt}
\caption{The rigidity test. Implied left-right hip separation $|\Delta z_{\mathrm{c}}|$ along the camera's depth axis, subscripted to distinguish it from the robot's vertical $z$, under a relative video model (VDA \cite{chen2025videodepth}, model units) and a single-image metric model (Depth Pro \cite{bochkovskii2025depthpro}, cm). CV is blank on the horse clips, where VDA leaves no separation to take a ratio of.
}
\label{tab:hipsep}
\begin{tabular*}{\linewidth}{@{\extracolsep{\fill}}lcccc@{}}
\toprule
& \multicolumn{2}{c}{VDA (relative)} & \multicolumn{2}{c}{Depth Pro (metric)} \\
\cmidrule(lr){2-3}\cmidrule(lr){4-5}
Clip & $|\Delta z_{\mathrm{c}}|$ & CV & $|\Delta z_{\mathrm{c}}|$ (cm) / CV & sign \\
\midrule
Dog walk   & $2.42$ & $40\%$  & 9 / $22\%$  & 98\% \\
Horse walk & $0.00$ & ---     & 6 / $35\%$  & 86\% \\
Horse trot & $0.01$ & ---     & 4 / $58\%$  & 61\% \\
Dog bow    & $1.68$ & $75\%$  & 11 / $28\%$ & 95\% \\
Give-paw   & $0.94$ & $107\%$ & 10 / $31\%$ & 94\% \\
\bottomrule
\end{tabular*}
\end{table}

\textbf{The rigidity test.} 
A depth model can also be tested with no training and no ground truth at all. The left-right hip separation it implies is a rigid distance, so a faithful model must return it near-constant. We run the two strongest current families, a temporally consistent relative video model, Video Depth Anything (VDA) \cite{chen2025videodepth}, and a single-image metric model, Depth Pro \cite{bochkovskii2025depthpro}, both recent and widely used. For each we measure the implied separation $|\Delta z_{\mathrm{c}}|$, its coefficient of variation (CV) over the clip, and the fraction of frames matching the modal left-right ordering.

Neither model holds the separation constant (\autoref{tab:hipsep}).
VDA varies by $40-107\%$ of its own mean on the dog clips and returns nearly zero separation on both horse clips, so on a horse it reports that the two hips are in the same place. Depth Pro roughly halves the swing, but its CV still reaches tens of percent, and on the trot it puts the left and right hip on the wrong side of each other in two frames out of five, barely above the $50\%$ a coin would give. On the dog walk that is about $2$ cm of frame-to-frame drift on a quantity that cannot move, set against the $4.5$ cm peak-to-peak of true paw travel the motion itself contains (\autoref{subsec:exp-mocap}). The same pattern holds across all $14$ screened clips, where Depth Pro's CV has a median of $33\%$.

\draft{
\textbf{Evaluating against independent targets.} Because $e_q^{\mathrm{ref}}$ is measured against the reference the pinned pipeline itself trained on, we re-evaluate every variant against two targets that none of them was trained on. The first reprojects each policy's motion into the source image plane and evaluates it against the detected 2D keypoints, which depends on no reference at all. It gives $5.8\%$ of the animal's bounding-box diagonal for the pinned policy, $6.9\%$ for repaired depth and $14-22\%$ for the variants that collapsed. The second evaluates every variant against the MoCap reference of \autoref{subsec:exp-mocap} on the rendered dog walk, ground truth that none of them sees, and gives $15.0^\circ$, $17.2^\circ$ and $28^\circ$. Both preserve the ordering of \autoref{fig:abl}.

\textbf{An articulated reconstruction passes the test.} The drift result is about per-pixel depth. Run on our clips, the articulated differentiable-rendering reconstruction that SLoMo \cite{zhang2023slomo} builds on \cite{yang2023lab4d} holds the hip separation to a $3.4\%$ CV. Its errors are in global scale and articulation instead, with front paws $66$ cm apart on a $60$ cm dog and feet reaching $1.46\times$ standing height against ANYmal-D's $1.18$ (under our scaling, not the authors' numbers). A constant stance width avoids that geometric mismatch, not jitter. Against a well-configured learned reconstruction our contribution is cost and per-clip robustness rather than fidelity (\autoref{subsec:exp-baselines}).

\textbf{Sweeping the pin, and an anatomical alternative.} Retraining the dog walk at $y^{\text{nom}}\in\{0,\,0.5,\,1,\,1.5\}\times w/2$, the three non-degenerate settings fall within $1.7^\circ$ of one another ($14.3-16.0^\circ$). The midline setting falls ($19.5^\circ$) because a zero-width stance has no roll moment arm. We also replace the constant with a species-level anatomical prior, the average lateral paw profile of \cite{zhang2018modeadaptive}'s dog MoCap replayed open-loop. This template \textit{knows} the true excursion, and it lands at $14.1^\circ$ against the constant's $14.3^\circ$. At this fidelity a constant and a correct species-level prior are indistinguishable.

\textbf{The ablation isolates one axis.} Inserting the video-depth model everywhere \textit{except} the pinned axis leaves the reference unchanged (median per-frame joint difference $0.00^\circ$, and RMS below $0.28^\circ$ on four of five behaviors, $1.4^\circ$ on the bow from a heading-estimate branch flip on a few frames). The ablations therefore isolate the lateral axis and nothing else. The $+$depth trot policy's $15.7^\circ$ in \autoref{fig:abl} is PPO variance, not a real gap. On retraining, the two variants give $11.9\,[11.3,12.5]^\circ$ (full) and $12.4\,[11.7,13.1]^\circ$ ($+$depth), with overlapping intervals.
}

\subsection{Is Root Recovery Necessary?}\label{subsec:exp-root}
\noindent 
To isolate the root-recovery step of \autoref{subsec:retarget}, we remove it and repeat the same procedure. Without it the reference carries no forward speed ($\hat v_x{=}0$), and everything else is unchanged.
This variant is the gray ``-root recovery'' bar of \autoref{fig:abl}. The policy matches the leg motion and never travels. It also gives the lowest $e_q^{\mathrm{ref}}$ of any variant on both clips, because a robot that stays where it is has nothing to fall out of phase with. Root recovery is therefore what converts a matched leg motion into locomotion, and the bar is a reminder that $e_q$ is only readable next to speed and survival. 

\draft{
\textbf{Hip regularization and the phase clock are conveniences.} 
To understand the effect of hip regularization, we retarget the dog walk and the horse trot with $\lambda_h{=}0$ instead of the default value of $5$.
This moves the reference by only $0.15-0.24^\circ$, and retraining from it leaves tracking unchanged: the $-$hip reg.\ bars of \autoref{fig:abl} land within one seed standard deviation of the full method on both clips ($15.9$ against $14.3^\circ$ on the walk, $11.7$ against $11.6^\circ$ on the trot, for a $\pm1.3-1.4^\circ$ seed spread), so the two are indistinguishable at this resolution. The phase clock is similarly uncritical: a per-stride clock is equal to or marginally better than our whole-segment clock ($13.8$ and $11.4^\circ$ against $14.3$ and $11.6^\circ$), and removing it costs under $1.3^\circ$. Unlike root recovery, neither stage is load-bearing. 
}

\subsection{What Video Costs Against MoCap}\label{subsec:exp-mocap}
\noindent
MoCap measures the true lateral motion that our pipeline discards, so it gives the one setting in which the video route can be evaluated against ground truth. From~\cite{zhang2018modeadaptive} we take dog MoCap comprising $5$ walk, $4$ trot and $3$ canter sequences. Two routes then run on the \textit{same underlying motion}. The \textit{MoCap route} retargets the 3D markers directly into reference joint trajectories $\hat{\mathbf{q}}^t$. The \textit{video route} renders that motion to a side-view video, discards the 3D entirely, and recovers a reference from the pixels alone with \ourMethod. Both use the same retargeter and the same reward, so any difference between them is exactly what the video pipeline loses.

The comparison is falsifiable in the direction that matters, since a pinned axis that diverged more than the sagittal joints would sink the central claim. Measured between the references, before any policy is trained, the sagittal disagreement is $12.0^\circ$ against $4.0^\circ$ on the pinned lateral joints, and a perfect tracker would inherit both. Normalized, the two readings disagree. The pinned hip abduction sweeps about $6^\circ$ over a stride in the deployed dog-walk reference against $53^\circ$ for the sagittal joints, so the errors are $70\%$ and $23\%$ of range and the ordering reverses. We therefore rest the claim on metric scale.

\begin{table}[t]
\centering
\caption{What video costs against MoCap, on one dog walk taken from markers and from a side-view rendering of the same motion. Both routes share the retargeter, reward and budget, and both are evaluated against the marker-derived reference.}
\label{tab:mocap}
\begin{tabular}{@{}p{0.4\columnwidth}cc@{}}
\toprule
& MoCap route & Video route (ours) \\
\midrule
$e_q$ (deg)$\downarrow$        & $\pmv{9.5}{1.0}$  & $\pmv{12.5}{1.4}$ \\
$e_p$ (cm)$\downarrow$         & $\pmv{10.5}{1.5}$ & $\pmv{13.5}{1.8}$ \\
Surv.\ (s, cap 20)              & 20.0        & 20.0 \\
\bottomrule
\end{tabular}
\end{table}

Trained end to end, the two routes differ by $3.0^\circ$ and $3.0$ cm and both survive the horizon (\autoref{tab:mocap}). That gap is the price of video at this tracking bandwidth, on a rendered clip with no blur, occlusion or compression, so it is a lower bound on the cost of real footage.

\textbf{The scale that settles it.} The true peak-to-peak lateral paw excursion of this walk is $4.5$ cm, a mean deviation near $1.4$ cm, while the policies already leave $3.5$ cm of lateral foot error (\autoref{tab:all}). Perfect lateral tracking would therefore remove at most $1.4$ of those $3.5$ cm, under a tenth of the total foot error, and unlike the angular split that comparison needs no normalizer. Across the twelve sequences the excursion is $\pmv{4.7}{1.2}$ cm for walks and $\pmv{6.0}{1.5}$ cm for trots, but $\pmv{11.0}{3.0}$ cm for canters, comparable to the foot error rather than small against it, so the design is bounded to symmetric gaits.

\draft{
\textbf{The value of the lateral axis.} The true peak-to-peak lateral paw excursion of this walk is $4.5$ cm, or a mean absolute deviation of about $1.4$ cm for a roughly sinusoidal profile. The policies already leave over $3.5$ cm of lateral foot error on the pinned axis, and roughly three times that fore-aft (\autoref{tab:all}). Tracking the true lateral motion perfectly would therefore remove at most $1.4$ of those $3.5$ cm, about a tenth of the total foot error, and the $3.0^\circ$ advantage of MoCap over video comes mostly from better sagittal geometry. A fourth configuration isolates the axis directly. A reference with MoCap's sagittal joints but our pinned lateral coordinate falls within seed noise of the full MoCap route on both metrics (\autoref{tab:mocap}). Even with sagittal geometry that video does not yet provide, restoring the true lateral motion is worth under half a degree at this tracking bandwidth.

\textbf{Where the conclusion reverses, and where it breaks.} A better reference does improve tracking, so the pipeline is not at a floor. Recovering the lateral axis becomes the largest remaining reference-level improvement once the sagittal disagreement falls from today's $12.0^\circ$ to the $4.0^\circ$ the pinned axis contributes, a $3\times$ gain in sagittal fidelity. Until then, keypoint quality on the axes the camera sees is what limits the video route. The rendered clip is easier than internet footage, with no blur, occlusion or compression, so this cost of video is a lower bound. The $4.5$ cm comes from a single sequence, so we also measure the excursion across twelve sequences of the dataset spanning three gaits (\autoref{tab:mocap}). Walks and trots confirm the regime, while canters reach $11.0\pm3.0$ cm, comparable to the lateral foot error itself, so the design should not be applied to asymmetric gaits without recovering the axis. A synthetic lateral oscillation injected into the dog-walk keypoints shows the failure mode, in which the pipeline silently drops the motion and yields a clean, trainable gait that is the wrong behavior. These excursion figures are for dogs. For horses we rely on the parasagittal premise and on \autoref{tab:hipsep}.

}

\subsection{Other Ablations}\label{subsec:exp-abl}
\noindent
\textbf{The shared weight vector.} The three per-clip reward scalars can be replaced by one fixed vector ($w_r{=}0.15$, $w_a{=}0.3$, $w_z{=}8$) for all five behaviors. Retrained that way, every behavior still survives the full horizon and stays far below the do-nothing floor (\textit{$e_q$ shared} in \autoref{tab:all}). It beats the per-clip scalars on both gestures ($4.2$ against $4.7^\circ$, $4.6$ against $8.3^\circ$), sits within one seed standard deviation on the trot, and is $2-4^\circ$ behind on the two walks. No behavior needs weights of its own, and the per-clip choice only sharpens the walks.


\draft{
\textbf{Camera roll} is the input assumption the pipeline tolerates least. 
On the dog walk clip, rotating the 2D keypoints in-plane and re-running the fixed pipeline end to end with no finetuning involved, drifts the reference by roughly $1.3^\circ$ of joint error per degree of roll, smoothly up to $20^\circ$, before IK branch flips break the reference at $30^\circ$. The drift is smooth but not small, since a $10^\circ$ roll already displaces the reference by $13.3^\circ$, more than the $10.0^\circ$ that separates our video reference from MoCap (\autoref{subsec:exp-mocap}). We do not correct for roll, so clips must be close to level\aref{app:retarget}.
}

\subsection{Does the Pipeline Hold Beyond the Five Clips?}\label{subsec:exp-breadth}
\noindent
To test that the five deployed behaviors are not a curated subset, we train the other nine screened clips with the shared weight vector and no per-clip choices of any kind. Seven of nine survive the full horizon on every seed and preserve their reference's contact timing, with the stride fraction separating a diagonal pair's touchdowns staying within $8$ percentage points of the reference on the gaits. Tracking error over the nine averages $16.1^\circ$ (range $12.8-21.7^\circ$) against $13.4-18.1^\circ$ for the three deployed gaits under the same vector, and we do not read differences inside that band. The two clips that fall on at least one seed carry the lowest mean keypoint confidence in the set, below $0.70$ with over $15\%$ of frames under $c_{\min}$, and both still locomote on their surviving seeds. Keypoint confidence is the best single predictor of tracking error, so the pipeline yields trainable, gait-faithful references well beyond the five we deploy, bounded by detector quality on the axes it uses.

\section{Conclusion and Limitations}\label{sec:limitations}
\noindent
\ourMethod turns a single uncalibrated side-view video into a quadrupedal skill on hardware, with a frozen 2D detector, a fixed analytic retargeting, and no depth estimate, per-clip optimization or motion capture. One reward structure learns walking and trotting across two species and two gestures, all zero-shot on ANYmal-D. We then measure, rather than assume, what the omitted axis was worth. A depth model cannot hold a rigid distance steady, and the true lateral motion is a fraction of the lateral foot error the policies already leave, so the depth stage can go at no cost to the controller.

The result is conditional in three ways. It assumes sagittal locomotion under a roughly static camera, which two in five casual clips fail. It is bounded to symmetric gaits, since a canter's lateral excursion is comparable to the foot error itself. And the rigidity argument covers per-pixel monocular depth, not articulated 3D reconstruction, which passes it and against which we compare on cost. The clearest next step is better sagittal keypoints, the current bottleneck, then combining the per-clip policies, either as one multi-skill policy or under a skill-selection layer over the pretrained ones \cite{wu2025vocaloco}, and a second platform.

\bibliographystyle{IEEEtran}
\bibliography{references}

@inproceedings{videomimic,
  title     = {Visual imitation enables contextual humanoid control},
  author    = {Allshire, Arthur and Choi, Hongsuk and Zhang, Junyi and McAllister, David 
               and Zhang, Anthony and Kim, Chung Min and Darrell, Trevor and Abbeel, 
               Pieter and Malik, Jitendra and Kanazawa, Angjoo},
  booktitle = {Conference on Robot Learning},
  year      = {2025}
}

@inproceedings{he2025asapaligningsimulationrealworld,
      title={ASAP: Aligning Simulation and Real-World Physics for Learning Agile Humanoid Whole-Body Skills}, 
      author={Tairan He and Jiawei Gao and Wenli Xiao and Yuanhang Zhang and Zi Wang and Jiashun Wang and Zhengyi Luo and Guanqi He and Nikhil Sobanbab and Chaoyi Pan and Zeji Yi and Guannan Qu and Kris Kitani and Jessica Hodgins and Linxi Fan and Yuke Zhu and Changliu Liu and Guanya Shi},
      year={2025},
      booktitle={Robotics: Science and Systems}
}

@article{mittal2025isaac,
  title={Isaac lab: A gpu-accelerated simulation framework for multi-modal robot learning},
  author={Mittal, Mayank and Roth, Pascal and Tigue, James and Richard, Antoine and Zhang, Octi and Du, Peter and Serrano-Munoz, Antonio and Yao, Xinjie and Zurbr{\"u}gg, Ren{\'e} and Rudin, Nikita and others},
  journal={arXiv preprint arXiv:2511.04831},
  year={2025}
}

@misc{schulman2017ppo,
      title={Proximal Policy Optimization Algorithms},
      author={John Schulman and Filip Wolski and Prafulla Dhariwal and Alec Radford and Oleg Klimov},
      year={2017},
      eprint={1707.06347},
      primaryClass={cs.LG},
      journal={arXiv preprint arXiv:1707.06347},
}

@article{zhang2023slomo,
  title={Slomo: A general system for legged robot motion imitation from casual videos},
  author={Zhang, John Z and Yang, Shuo and Yang, Gengshan and Bishop, Arun L and Gurumurthy, Swaminathan and Ramanan, Deva and Manchester, Zachary},
  journal={IEEE Robotics and Automation Letters},
  volume={8},
  number={11},
  pages={7154--7161},
  year={2023},
  publisher={IEEE}
}

@inproceedings{yao2022imitation,
  title={Imitation and adaptation based on consistency: A quadruped robot imitates animals from videos using deep reinforcement learning},
  author={Yao, Qingfeng and Wang, Jilong and Yang, Shuyu and Wang, Cong and Zhang, Hongyin and Zhang, Qifeng and Wang, Donglin},
  booktitle={IEEE International Conference on Robotics and Biomimetics},
  pages={1414-1419},
  year={2022}
}

@inproceedings{bochkovskii2025depthpro,
  author     = {Aleksei Bochkovskii and Ama\"{e}l Delaunoy and Hugo Germain and Marcel Santos and
               Yichao Zhou and Stephan R. Richter and Vladlen Koltun},
  title      = {Depth Pro: Sharp Monocular Metric Depth in Less Than a Second},
  booktitle  = {International Conference on Learning Representations},
  year       = {2025},
}

@article{zhang2018modeadaptive,
author = {Zhang, He and Starke, Sebastian and Komura, Taku and Saito, Jun},
title = {Mode-adaptive neural networks for quadruped motion control},
year = {2018},
volume = {37},
number = {4},
journal = {ACM Trans. Graph.},
month = jul,
articleno = {145},
numpages = {11},
}

@article{chane2024reinforcement,
  title={Reinforcement learning from wild animal videos},
  author={Chane-Sane, Elliot and Roux, Constant and Stasse, Olivier and Mansard, Nicolas},
  journal={arXiv preprint arXiv:2412.04273},
  year={2024}
}

@article{peng2021amp,
  title={Amp: Adversarial motion priors for stylized physics-based character control},
  author={Peng, Xue Bin and Ma, Ze and Abbeel, Pieter and Levine, Sergey and Kanazawa, Angjoo},
  journal={ACM Transactions on Graphics},
  volume={40},
  number={4},
  pages={1--20},
  year={2021},
  publisher={ACM New York, NY, USA}
}

@article{wu2025vocaloco,
  title={VOCALoco: Viability-Optimized Cost-Aware Adaptive Locomotion},
  author={Wu, Stanley and Danesh, Mohamad H and Li, Simon and Yurchyk, Hanna and Abyaneh, Amin and El Houssaini, Anas and Meger, David and Lin, Hsiu-Chin},
  journal={IEEE Robotics and Automation Letters},
  volume={11},
  number={2},
  pages={1146--1153},
  year={2025},
  publisher={IEEE}
}

@inproceedings{chen2025videodepth,
  title={Video depth anything: Consistent depth estimation for super-long videos},
  author={Chen, Sili and Guo, Hengkai and Zhu, Shengnan and Zhang, Feihu and Huang, Zilong and Feng, Jiashi and Kang, Bingyi},
  booktitle={Proceedings of the Computer Vision and Pattern Recognition Conference},
  pages={22831-22840},
  year={2025}
}

@inproceedings{joska2021acinoset,
  title={Acinoset: A 3d pose estimation dataset and baseline models for cheetahs in the wild},
  author={Joska, Daniel and Clark, Liam and Muramatsu, Naoya and Jericevich, Ricardo and Nicolls, Fred and Mathis, Alexander and Mathis, Mackenzie W and Patel, Amir},
  booktitle={IEEE International Conference on Robotics and Automation},
  pages={13901-13908},
  year={2021}
}

@inproceedings{rueegg2023bite,
  title={Bite: Beyond priors for improved three-d dog pose estimation},
  author={R{\"u}egg, Nadine and Tripathi, Shashank and Schindler, Konrad and Black, Michael J and Zuffi, Silvia},
  booktitle={Proceedings of the IEEE/CVF Conference on Computer Vision and Pattern Recognition},
  pages={8867--8876},
  year={2023}
}

@article{ye2024superanimal,
  title={SuperAnimal pretrained pose estimation models for behavioral analysis},
  author={Ye, Shaokai and Filippova, Anastasiia and Lauer, Jessy and Schneider, Steffen and Vidal, Maxime and Qiu, Tian and Mathis, Alexander and Mathis, Mackenzie Weygandt},
  journal={Nature communications},
  volume={15},
  number={1},
  pages={5165},
  year={2024},
  publisher={Nature Publishing Group UK London}
}

@article{danesh2026toward,
  title={Toward Hardware-Agnostic Quadrupedal World Models via Morphology Conditioning},
  author={Danesh, Mohamad H and Li, Chenhao and Abyaneh, Amin and Houssaini, Anas and Ellis, Kirsty and Berseth, Glen and Hutter, Marco and Lin, Hsiu-Chin},
  journal={arXiv preprint arXiv:2604.08780},
  year={2026}
}

@article{danesh2025safe,
  title={Safe domain randomization via uncertainty-aware out-of-distribution detection and policy adaptation},
  author={Danesh, Mohamad H and Wabartha, Maxime and Wu, Stanley and Pineau, Joelle and Lin, Hsiu-Chin},
  journal={arXiv preprint arXiv:2507.06111},
  year={2025}
}

@inproceedings{yang2023lab4d,
  title={Reconstructing animatable categories from videos},
  author={Yang, Gengshan and Wang, Chaoyang and Reddy, N Dinesh and Ramanan, Deva},
  booktitle={IEEE/CVF Conference on Computer Vision and Pattern Recognition},
  pages={16995-17005},
  year={2023}
}

@article{fischer2002basic,
  title={Basic limb kinematics of small therian mammals},
  author={Fischer, Martin S and Schilling, Nadja and Schmidt, Manuela and Haarhaus, Dieter and Witte, Hartmut},
  journal={Journal of Experimental Biology},
  volume={205},
  number={9},
  pages={1315-1338},
  year={2002},
  publisher={Company of Biologists}
}

@inproceedings{peng2020learning,
  title={Learning agile robotic locomotion skills by imitating animals},
  author={Peng, Xue Bin and Coumans, Erwin and Zhang, Tingnan and Lee, Tsang-Wei and Tan, Jie and Levine, Sergey},
  booktitle={Robotics: Science and Systems},
  year={2020}
}

@article{hildebrand1989quadrupedal,
  title={The quadrupedal gaits of vertebrates.},
  author={Hildebrand, Milton},
  journal={Bioscience},
  volume={39},
  number={11},
  pages={766},
  year={1989}
}

@article{peng2018deepmimic,
  title={Deepmimic: Example-guided deep reinforcement learning of physics-based character skills},
  author={Peng, Xue Bin and Abbeel, Pieter and Levine, Sergey and Van de Panne, Michiel},
  journal={ACM Transactions On Graphics},
  volume={37},
  number={4},
  pages={1--14},
  year={2018},
  publisher={ACM New York, NY, USA}
}

@inproceedings{margolisyang2022rapid,
  title={Rapid Locomotion via Reinforcement Learning},
  author={Margolis, Gabriel and Yang, Ge and Paigwar, 
          Kartik and Chen, Tao and Agrawal, Pulkit},
  booktitle={Robotics: Science and Systems},
  year={2022}
}

@article{lee2020learning,
  title={Learning quadrupedal locomotion over challenging terrain},
  author={Lee, Joonho and Hwangbo, Jemin and Wellhausen, Lorenz and Koltun, Vladlen and Hutter, Marco},
  journal={Science robotics},
  volume={5},
  number={47},
  pages={eabc5986},
  year={2020}
}

@inproceedings{rudin2022learning,
  title={Learning to walk in minutes using massively parallel deep reinforcement learning},
  author={Rudin, Nikita and Hoeller, David and Reist, Philipp and Hutter, Marco},
  booktitle={Conference on Robot Learning},
  pages={91--100},
  year={2022},
  organization={PMLR}
}

@ARTICLE{yoon2024stmr,
  author={Yoon, Taerim and Kang, Dongho and Kim, Seungmin and Cheng, Jin and Ahn, Min Sung and Coros, Stelian and Choi, Sungjoon},
  journal={IEEE Transactions on Robotics}, 
  title={Spatio-Temporal Motion Retargeting for Quadruped Robots}, 
  year={2025},
  volume={41},
  number={},
  pages={5471-5490},
  doi={10.1109/TRO.2025.3600123}}

@article{zhao2025learning,
  title={Learning aggressive animal locomotion skills for quadrupedal robots solely from monocular videos},
  author={Zhao, Liu and Luo, Zeren and Han, Yimin and Zhang, Jiahui and Chen, Yuanhao and Liu, Yunhui and Lu, Peng},
  journal={npj Robotics},
  volume={3},
  number={1},
  pages={32},
  year={2025},
  publisher={Nature Publishing Group UK London}
}

@article{mathis2018deeplabcut,
    title = {DeepLabCut: markerless pose estimation of user-defined body parts with deep learning},
    author = {Alexander Mathis and Pranav Mamidanna and Kevin M. Cury and Taiga Abe  and Venkatesh N. Murthy and Mackenzie W. Mathis and Matthias Bethge},
    journal = {Nature Neuroscience},
    year = {2018},
    }

@inproceedings{gurram2024rl,
      title={Reinforcement Learning For Quadrupedal Locomotion: Current Advancements And Future Perspectives}, 
      author={Maurya Gurram and Prakash Kumar Uttam and Shantipal S. Ohol},
      year={2025},
    booktitle={International Conference on Mechanical Engineering and Robotics Research},
}

@article{han2024lifelike,
  title={Lifelike agility and play in quadrupedal robots using reinforcement learning and generative pre-trained models},
  author={Han, Lei and Zhu, Qingxu and Sheng, Jiapeng and Zhang, Chong and Li, Tingguang and Zhang, Yizheng and Zhang, He and Liu, Yuzhen and Zhou, Cheng and Zhao, Rui and others},
  journal={Nature Machine Intelligence},
  volume={6},
  number={7},
  pages={787-798},
  year={2024},
  publisher={Nature Publishing Group UK London}
}

@article{alexander1983dynamic,
  title={A dynamic similarity hypothesis for the gaits of quadrupedal mammals},
  author={Alexander, R McN and Jayes, AS},
  journal={Journal of zoology},
  volume={201},
  number={1},
  pages={135--152},
  year={1983},
  publisher={Wiley Online Library}
}

\end{document}